# Semi-supervised Clustering Ensemble by Voting

Ashraf Mohammed Iqbal[1], Abidalrahman Moh'd[2], and Zahoor Ali Khan[3]
Faculty of Computer Science[1,3], Faculty of Engineering Mathematics & Internetworking[2]
Dalhousie University, Halifax, Canada.
{iqbal[1], zkhan[3]}@cs.dal.ca, Abidalrahman.Mohd@dal.ca[2]

*Abstract*— **Clustering ensemble is one of the most recent advances in unsupervised learning. It aims to combine the clustering results obtained using different algorithms or from different runs of the same clustering algorithm for the same data set, this is accomplished using on a consensus function, the efficiency and accuracy of this method has been proven in many works in literature.**

**In the first part of this paper we make a comparison among current approaches to clustering ensemble in literature. All of these approaches consist of two main steps: the ensemble generation and consensus function. In the second part of the paper, we suggest engaging supervision in the clustering ensemble procedure to get more enhancements on the clustering results. Supervision can be applied in two places: either by using semi-supervised algorithms in the clustering ensemble generation step or in the form of a feedback used by the consensus function stage.**

**Also, we introduce a flexible two parameter weighting mechanism, the first parameter describes the compatibility between the datasets under study and the semi-supervised clustering algorithms used to generate the base partitions, the second parameter is used to provide the user feedback on the these partitions. The two parameters are engaged in a "relabeling and voting" based consensus function to produce the final clustering.**

## I. INTRODUCTION

RECENT research on unsupervised learning is increasingly focusing on combining multiple data partitions as a way to improve the accuracy of clustering outcomes. Many clustering algorithms are capable of producing different results for the same data; these results capture various distinct aspects of the data and provide improved overall clustering than what is typically achieved by a single clustering algorithm. It has been shown that a meaningful consensus of multiple clusterings is possible by using a consensus function that maps a given ensemble (a collection of different partitions of a data set) to a combined clustering result.

A variety of efficient consensus functions have been proposed so far, based on statistical, graph based or information theoretic approaches. [1, 2] propose a consensus functions that is based on voting, co-association matrix is used in [3, 4, 5], and a hyper graph cuts approach in [6, 7], mixture models in [8] and based on mutual information in [9]. Extensive experiments with these functions indicate that a combination of clustering is capable of detecting novel cluster structures. Empirical evidence also supports the idea that requirements for individual clustering algorithms can be significantly relaxed in favor of weaker and inexpensive partition generation.

Semi-supervised clustering is another hot topic in machine learning, in which both labeled and unlabeled data are used for training - typically a small amount of labeled data with a large amount of unlabeled data. Semi-supervised learning falls between supervised learning (with completely labeled training data) and unsupervised learning (without any labeled training data). It has been proven by many machine-learning researchers that unlabeled data, when used in conjunction with a small amount of labeled data, can produce considerable improvement in learning accuracy. Generating labeled data for a learning problem often requires a human effort (supervision) to manually classify training examples. The cost producing a fully labeled training set in the labeling process may be infeasible in many cases, whereas producing unlabeled data is relatively inexpensive. In such situations, semi-supervised learning can be of great practical value.

Many semi-supervised algorithms were proposed in literature with various methodologies, some based on EM with generative mixture models, self-training, co-training, Transductive Support Vector Machines (TSVM), and graph-based methods. Because labeled data is scarce, semi-supervised learning methods make strong model assumptions. Ideally we should use a method whose assumptions fit the problem structure. This may be difficult in reality. Generally, EM with generative mixture models may be a good choice if the classes produce well clustered data; co-training may be appropriate when the features naturally split into two sets; graph-based methods can be used if points with similar features tend to be in the same class. But there is no direct way for choosing the type of semi-supervised algorithm.

In this paper, we aim to combine the benefit of clustering ensembles' and semi-supervised clustering algorithms to improve clustering accuracy for the results, supervision can be provided in two places: either by using semi-supervised algorithms in the clustering ensemble generation step or in the form of a feedback used in the consensus function stage. We also use two flexible weighting techniques to give the supervisor the ability to tune the clustering process towards the clustering algorithm that fits the type of the input data sets and to provide a feedback after the clustering generation step to a "relabeling and voting" based consensus function. These parameters facilitate biasing the clustering results to the correct direction.

In the following section we will make a comparison among different approaches to clustering ensemble. In section III we explain our proposed semi-supervised clustering ensemble algorithm that is based on relabeling and voting consensus function, and then we provide the future work and conclude in section IV.

## II. CLUSTERING ENSEMBLE

The objective of clustering Ensemble is to integrate clustering partitions obtained using various techniques. Clustering ensemble algorithms are usually divided into two stages. At the first stage different partitions of same dataset are produced using independent runs of various clustering algorithms or the same clustering algorithms. Then, in the next stage, a consensus function is used to find a final partition from the partitions generated in the first stage. Figure 1 shows clustering ensembles procedure.

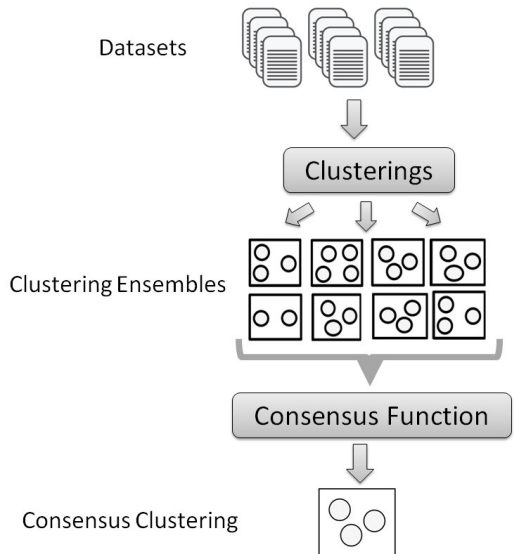

Figure 1: Overview of clustering ensemble procedure.

There are various approaches of consensus function: Mixture Models - a probabilistic model of consensus using a finite mixture of multinomial distributions in a space of clusterings, Voting Based Approach - solving explicitly the correspondence problem between the labels of known and derived clusters [17] and determining the final consensus function by assigning objects in clusters by using a simple voting procedure, Information Theory Approach – considering the Mutual Information between the probability distribution of labels of the consensus partition and the labels of the ensemble as the objective function for ensemble, Co-Association Based Approach - considering the number of clusters shared by two objects in all the partition of an ensemble as the measurement of similarity between those two objects, Hyper Graph Based Approach – transforming the initial set of clusters into a hyper-graph representation and solving the minimum-cut hyper-graph partitioning problem.

We will compare the different consensus methods discussed above in terms of computational complexity, scalability, robustness, and ease to implement the algorithms. The overall comparison is summarized in Table 1.

| Consensus Function | Computational Complexity | Scalability | Robustness | Ease of Implementation |
|---|---|---|---|---|
| Mixture Models | $O(k^3)$ [8], [13] | High | Low robustness compared to some other consensus functions | Easy to implement |
| Voting Based Approach | $O(k^3)$ [1], [8], [15]; however, Ayad and Kamel [17] proposed one with complexity linear to $N$. | High | High | Easy to implement |
| Information Theory Approach | Generally $O(k^3)$ [18]; Topchy et al. [9] proposed one with O(kHN) | Low scalability compared to other methods. | High | Not easy to implement many such methods [19] |
| Co-Association Based Approach | $O(N^2)$ | High | High | Difficult to implement |
| Hyper-graph Based Approach | Generally $O(N^3)$ [14]; Fern and Brodley [11] proposed one with O(kN) | High | High | Difficult to implement |

Table 1: Comparison among different approaches of consensus function.

Many models have been proposed on Mixture Model based consensus function. Some models [8], [10], [13] used EM algorithm for solving the maximum likelihood problem, i.e., finding a combined partition, whereas other [12] used genetic algorithm for the same purpose. These methods are comprehensible with high scalability, easy to implement, but were computationally more expensive than some other consensus methods. Generally, the complexity is $O(k^3)$ [8], [13] whereas it is $O(N^2)$ for the one using genetic algorithm [12].

The simplest consensus functions are voting based approaches. These methods are straight forward and easy to implement, provide high robustness and high scalability. Although the proposed methods using voting approach are computationally more expensive (withy computational complexity $O(k^3)$ [1], [8], [15], Ayad and Kamel [16] proposed one probabilistic method of voting with complexity linear to $N$. They could also solve the problem of knowing the number of final partitions in advance and could provide one solution for random number of such partitions.

Comparatively high robustness could be achieved by using consensus functions based on information theory where the mutual information between the probability distribution of labels of the consensus partition and the labels of the ensemble were tried to maximize. Generally, the computational complexity was high ($O(k^3)$ [18]). However, Topchy et al. [9] proposed one method with complexity linear to N, O(kHN). The algorithms were difficult to implement as it might require a few restarts of these algorithms in case of low local minima.

Co-association based consensus functions also provide comparatively high robust clusters while some of these have the property of fast emergence. The implementations of such algorithms proved to be difficult because of employing hierarchical agglomerative clustering algorithms while combining the partitions. Also, all of proposed methods based on co-association matrices are computationally very much expensive with complexity $O(N^2)$.

The most computationally efficient consensus function was proposed by Fern and Brodley [11] based on hyper-graph partitioning technique with complexity O (kN). Generally, the algorithms based on such technique provide high robustness where some are more robust than others. For an example, HBGF [11] algorithm could provide more robustness than both METIS [6] and SPEC [20]. Although Fern and Brodley [11] could introduce one algorithm which was computationally very much efficient, the other algorithms based on hypergraph approach were very expensive (with complexity $O(N^3)$). Moreover, these algorithms are difficult to implement because of lacking of simplicity.

III. SEMI-SUPERVISED CLUSTERING ENSEMBLE BY VOTING

As mentioned in the introduction the aim of this paper is to combine the benefit of clustering ensemble and semi-supervised clustering algorithms to improve clustering accuracy for the results, this is accomplished by engaging supervision in the clustering ensemble procedure in two places: either in the clustering ensemble generation step or in the form of a feedback on the generated partition that can be used in the consensus function stage. Figure 2 shows the places where supervision and feedback can be applied.

Figure 3 describes the proposed algorithm in detail, in the first stage different semi-supervised algorithms are used to generate the base partition, three aspects of benefit are provided at this stage, different types of supervision provide more flexibility by giving the user the choice of selecting the easiest and more efficient way from multiple types of supervision. Also multiple semi-supervised algorithms address different types supervision (seeding, labeling, constraint, weighting), and the different types of the unsupervised algorithms that the semi-supervised versions are based on address different aspects of the datasets under study.

In addition to the inherited supervision in the semi-supervised algorithms used in the first stage, two weighting parameters are incorporated; the main target of weighting is to efficiently incorporate supervision and user feedback about semi-supervised clustering algorithms as well as the base partitions to the consensus function.

The weighting first parameter $\alpha$ is used to give weighting to the semi-supervised algorithms used in the first stage, the user could give more weight to the algorithm that has more efficient *type of supervision* (e.g. constraints vs. labeling, etc.), or it can be assigned based on specific *aspect of functionality provided by the semi-supervised algorithm* (i.e. seeded K-Means is more efficient for noise, constraint and COP-Means are more efficient if the data is noise free [21], SPK-Means is more efficient for high dimensional data [22], etc.).

The second parameter $\beta$ is used to carry feedback about the first stage (base partitions) to the second stage (consensus function). Feedback can provide indicators about the good clustering that the user desire, this may include the part information that is not addressed by supervision. If the feedback and the supervision are consistent, this gives an indication of good accuracy for the results. If not, the user has to consider one of them more than the other, the amount of trustworthy between the supervision and the feedback is case dependent and we left it to the user to decide the amount of significance that each one of them must contribute in the final results.

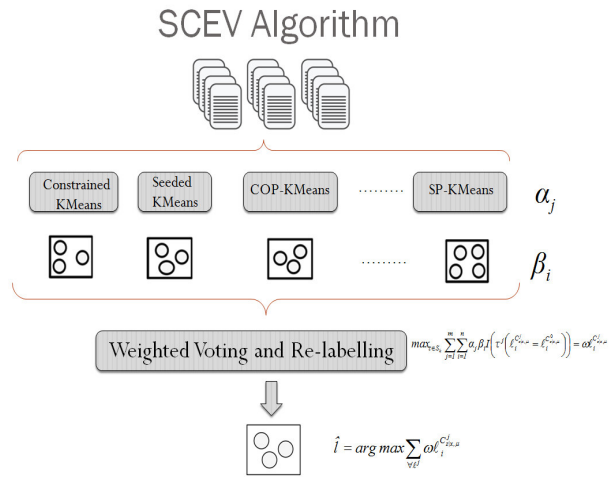

Figure 2: The places where supervision and feedback can be applied.

In the proposed algorithm voting and relabeling approach was selected for finding a consensus partition from the partitions generated by a variety of semi-supervised clustering algorithms. This approach was selected because of its simplicity and efficiency. Because it attempts to explicitly solve the correspondence problem between the labels of known and derived clusters. The final consensus function assigns objects to clusters by using a weighted voting procedure.

The relabeling procedure is started according to a fixed reference partition which can be selected by the supervisor (as a good partition) or it can be selected randomly from the base partitions if the supervisor doesn't have any hint about the quality of the partitions. Voting and relabeling approach is proved to be $O(k^3)$ [1], [8], this is much more efficient than other types of consensus functions because their complexity is a function of the number of object (data points), which is much larger than the number of partitions k [14].

Table 2 shows the original voting and labeling procedure, the columns in the tables represent the base partitions and the rows represent the objects (data points); a permutation procedure is used to align the object in a way that the maximum overlap between the similar clusters occur. To save computations, a reference partition is usually defined and the other partitions align to it.

|   | $C^1$ | $C^2$ | $C^3$ | $C^4$ |   | $C^1$ | $C^2$ | $C^3$ | $C^4$ |   | FC |
|---|---|---|---|---|---|---|---|---|---|---|---|
| $x_1$ | 1 | A | α | Z | $x_1$ | 1 | 1 | 1 | 2 |   | 1 |
| $x_2$ | 1 | A | β | Y | $x_2$ | 1 | 1 | 2 | 1 |   | 1 |
| $x_3$ | 3 | B | β | ? | $x_3$ | 3 | 3 | 2 | ? |   | 3 |
| $x_4$ | 2 | C | α | Y | $x_4$ | 2 | 2 | 1 | 1 |   | ? |
| $x_5$ | 2 | B | γ | Z | $x_5$ | 2 | 3 | 2 | 2 |   | 2 |
| $x_6$ | 3 | C | ? | Z | $x_6$ | 3 | 2 | ? | 2 |   | 2 |
| $x_7$ | 3 | B | γ | ? | $x_7$ | 3 | 3 | 2 | ? |   | 3 |

Table 2: Basic re-labelling and partitioning.

After permutation and alignment a relabeling for the objects is done, the objects that belong to similar clusters in different partitions are given the same label. The labeling for the final partitioning is decided by voting, the object is assigned the label that appears more in the relabeled base partitioning.

In the proposed algorithm a weighted version of re-labelling and voting procedure is used. The permutation to get the maximum overlap is done in the same way, but in the labelling process weights are assigned to the new labels. These weights are calculated by combining the Semi-supervised clustering algorithm weighting parameter $β_i$ and the partition feedback weighting parameter $α_j$.

The amount of contribution for each of these parameters depends on the degree of trustworthy of the feedback and the supervision which can be decided by the supervisor. The labeling for the final partitioning is decided by weighted voting; the object is assigned the label that have maximum sum of weights among the base partitioning.

Table 3: Weighted re-labeling and partitioning.

Semi-supervised Clustering Ensembles by Voting (SCEV) algorithm:

1. Apply the initial semi-supervised clustering algorithm $C^0_{z:x,u}$ to the learning date set $\chi$ to obtain cluster labels $\ell_i^{C^0_{z:x,u}}$

2. For each semi-supervised clustering algorithm $C_j$ do the following
   a. apply semi-supervised clustering algorithm $C_j$ to the learning set $\chi$ and obtain cluster labels
   b. Permute the cluster labels $\ell_i^{C^j_{z:x,u}}$ assigned to the learning set $\chi$ so that there is maximum overlap with the original clustering of these observations. More specially, let $S_k$ denote the set of all permutations of the integers $1, ..., K$. Find the permutation $τ \in S_k$ such that
   
   $$max_{τ \in S_k} \sum_{j=1}^{m} \sum_{i=1}^{n} α_j β_i I\left(τ^j\left(\ell_i^{C^j_{z:x,u}} = \ell_i^{C^0_{z:x,u}}\right)\right) = ω\ell_i^{C^j_{z:x,u}}$$
   
   - $I(.)$ Indicator function: equal to 1 if the condition in parentheses is true and 0 otherwise.
   - $τ(.)$ Permutation function: used to get matching with maximum overlap.
   - $β_i$ Semi-supervised clustering algorithm weighting parameter.
   - $α_j$ Partition feedback weighting parameter.

3. After assigning weighted cluster labels for each partition data points by maximum overlap, we calculate the winning partition label by majority vote.
   
   $$\hat{l} = arg max \sum_{\forall \ell^j} ω\ell_i^{C^j_{z:x,u}}$$

Figure 3: Proposed Semi-supervised Clustering Ensembles by Voting (SCEV) algorithm.

## IV. CONCLUSION

We critically compared among different approaches of consensus functions while considering different algorithms proposed under these approaches in the first part of this paper. In the second part, we proposed a novel semi-supervised clustering ensemble algorithm by incorporating supervision at ensemble generation step and user feedback to be used by consensus function for final partition generation. We introduced a flexible two parameters weighting mechanism in our proposed algorithm. The first parameter is used to describe the compatibility between the datasets under study and the semi-supervised clustering algorithms used to generate the base partitions where the second parameter is used to provide the user feedback on these base partitions. We used a relabeling and voting based consensus function to produce the target partition.

Combining supervision with clustering ensemble is assumed to give higher level of accuracy. The supervision at ensemble generation step can aid and bias different clustering to produce better and high quality base partitions. The consensus function can also take benefit from user feedback about the base partitions to produce higher quality target partition. This approach gives user flexibility of choosing multiple types of supervision and feedback in both steps. This type of weighting scheme can be applied to other consensus functions as well. However, we choose relabeling and voting based approach for its simplicity, high robustness, scalability and ease of implementation. The proposed consensus function is computationally very efficient with complexity linear to the data size; $N$.

Parallelism can be applied in both steps of clustering ensemble to further improve the complexity.

In future, we will implement our proposed algorithm, test it against multiple types of datasets, and compare with other clustering ensemble and semi-supervised algorithms. We will also try to incorporate Genetic Algorithm for voting and relabeling in the consensus function to produce better results.